# A Simple and Effective Approach for Fine Tuning Pre-trained Word Embeddings for Improved Text Classification

Amr Al-Khatib and Samhaa R. El-Beltagy

Center of Informatics Sciences,
Nile University, Giza, Egypt

`a.mehasseb@nu.edu.eg, samhaa@computer.org`

## Abstract

This work presents a new and simple approach for fine-tuning pretrained word embeddings for text classification tasks. In this approach, the class in which a term appears, acts as an additional contextual variable during the fine tuning process, and contributes to the final word vector for that term. As a result, words that are used distinctively within a particular class, will bear vectors that are closer to each other in the embedding space and will be more discriminative towards that class.

To validate this novel approach, it was applied to three Arabic and two English datasets that have been previously used for text classification tasks such as sentiment analysis and emotion detection. In the vast majority of cases, the results obtained using the proposed approach, improved considerably.

## Introduction

Over the past few years, deep learning has been being increasingly used in text classification tasks [1]. The success of deep learning for text classification, is usually highly dependent on the availability of large volumes of training data, and/or the existence of word embeddings that are relevant to the domain being targeted by the classification task. Word embeddings can be defined as the outcome of applying any set of techniques that result in words or phrases in some given vocabulary being mapped to vectors of real numbers. The motivation for this work arose from the need to make use of powerful deep learning models in cases where only small training datasets are available, and existing pretrained embeddings, are not producing desired outcomes.

Word embeddings, which are almost always a part of any neural based text classification system, arose to address the limitations of traditional text representation schemes. The conventional and most common method of representing text is through the bag of words (BOW) method [2] where any piece of text is represented by a vector, the length of which is the vocabulary encountered in some input corpus. However, since every word that has appeared in the input corpus, has an entry in the vector, representing any document or piece of text, is done through very sparse vectors. In addition, the BOW model, has no way of capturing semantic relationships between words. Semantic similarities can be important in classification and clustering. Consider for example two short documents; one stating the 'cupcakes' are 'yummy', and another stating that 'cakes' are 'delicious'. To a human reader, both these documents are very similar, but a BOW model may fail to identify this similarity as each of the words used in either document, will have its own independent entry in the BOW vector. The idea of representing individual words using dense vectors, which was introduced by word embedding models, addresses these limitations, as individual vectors act as semantic representatives of a word. In such models, the similarity between words, can be measured using the cosine similarity score between their vectors.

In the last few years, new neural network dependent methods for generating word embeddings (e.g. Word2Vec [3], and GloVe [4]) have emerged and have become quite popular. The neural network approach is based on training a neural network to predict a word vector from a context (neighboring words) or vice versa (predict the context from the word). The success of these models is dependent on the availability of large corpora which are used to train the model. The outcome of well trained models, are vectors in which semantically similar words tend to gather in the embedding space. These word vectors can be summed, averaged, or concatenated to provide a non-sparse representation for text.

While these approaches do tend to capture the semantics of individual words, the fact that words that appear in the same context have similar embeddings, can often result in undesirable effects with respect to specific tasks. Since the terms 'good' and 'bad' for example, often appear in similar contexts, they will usually appear close to each other in an embedding space. This might not be desirable for certain tasks such as sentiment analysis. For this specific task, allowing the class (positive, negative, and neutral) to act as an additional contextual factor that influences the final vector representation of words, can lead to improved classification results. This is the premise, on which this work is based.

The rest of this paper is organized as follows: Section 2 provides the needed background for this work, Section 3 overviews related work, Section 4 presents the proposed word embeddings fine tuning approach, Section 5 describes experiments that were carried out to evaluate the proposed approach, and Section 6 concludes this paper.

**Background**

Over the years, many different approaches and techniques have been proposed for building word embeddings. The ones that are particularly relevant to this work, are Word2Vec and Doc2Vec proposed by Mikolov et al. [3] and Le et al. [5] respectively. In Word2Vec, Mikolov et al. presented two new model architectures for continuous vector representation for a word. The two models have simple, shallow architectures and are relatively fast. The names given to the two models are: continuous bag of words (CBOW) and skip gram (SG). The speed of CBOW and SG provide a great capacity to learn word vectors from gigantic amounts of text (like millions of sentences and billions of words) in a relatively short period of time.

Doc2Vec is a similar approach that was introduces with the objective of learning vectors that represent whole sentences or documents and was inspired by the previously presented methods for learning word vectors [5]. To represent a whole sentence or document as a single vector, a vector for each sentence is created and used during the learning process. This basically means that a sentence impacts the numerical representation of each word within that sentence, in as much a way as the words that appear in that sentence impact the numerical representation of the sentence. In the training process, every sentence has a single unique vector in a matrix **D**, and every word also has its own unique vector in a matrix **W** where **D** represents all the unique sentence vectors, while **W** represents the matrix that combines all the word vectors. The sentence vector and all the vectors of the words in a window within that sentence are averaged or concatenated in order to predict the next word [5]. The window of words (the context) here is represented using a fixed window size that slides over the words in the document. The sentence vector is similar to the word vector, except that it retains the current sentence or document. It is important to highlight that word vectors (represented by the matrix **W**) are shared across all documents, so if there is a word that appears in two different documents, its numerical representation will be learned from both documents. Document vectors on the other hand, are learned from the words within that document only. Both the document vectors and word vectors are learned using stochastic gradient descent (SGD), where the gradients are computed by the backpropagation algorithm.

Word2Vec has been used for pretraining data from very large corpora for later use in deep learning models; Google News[1] and AraVec[6], are examples of two widely used pretrained embeddings for English and Arabic respectively. The use of these models for various tasks, can be considered a form of transfer learning [7].

---

[1] https://code.google.com/archive/p/word2vec/

**Related Work**

There is no doubt, that word embeddings are largely affected by training data that was used to generate them. These models are considered context free, as neither the task at hand, nor different semantic meanings that a word can carry depending on the context in which it appears, are taken into consideration. The recent introduction of Google's BERT model[2], primarily targets the latter point. Other efforts have been made to address this issue.

An example is the framework for supervised fine tuning of word embeddings proposed by Yang et al. [8]. This framework uses pretrained word embeddings in addition to lexical semantic resources. The framework has two phases, where the first phase consists of ranking data generation (for each word a ranking of semantically similar words is generated), and the second phase targets ranking fine tuning. The main idea of the proposed framework is to use word similarities ranking as the output of the training process. This framework is composed of two algorithms; the first algorithm handles the automated process of labeled data generation from the available word embeddings and knowledge resources, and the second is a prepared inverse error weighted minibatch stochastic gradient descent algorithm. The authors reported that their optimization algorithm can utilize the provided knowledge to finetune word embeddings without damaging the previous similarities between word vectors. The proposed framework's performance was evaluated on semantic similarity prediction and relational similarity tasks (intrinsic evaluation), and sentence completion and sentiment analysis tasks (extrinsic evaluation). The experiments were conducted on 10 datasets and with 6 different word embeddings, and the results indicate that the framework can improve the quality of word vectors.

Labutov et al. [9] presented a model to fine tune pretrained word embeddings for supervised tasks, where each sentence or document (e.g. movie review) was treated as a group of words, and this sentence was assigned a sentiment label of zero or one. The Conditional likelihood of the labels given the sentences was then computed and used for optimizing the embeddings through logistic regression. The IMDB dataset was used for experimentation and was split into 2 equal testing and training sets. The authors used randomly generated embeddings, and also vectors of zeros as well as two baseline embedding sets and trained them with the proposed model. In addition, three other pretrained word embedding sets were used as additional baselines. The reported results showed that the presented model has promising capabilities and successfully improved the results from the baselines.

In the work proposed by Gao et al. [10], a system of two models is used to adjust word embeddings to increase synonyms similarities and denoise word representations learned by word2Vec [3]. The first model is composed of stacked autoencoders that can get more compact and denoised word representations, where an autoencoder is trained for dimensionality reduction and the resulting weights and bias are fed to the second autoencoder which reconstructs the input. Then a model is used for fine-tuning which optimizes the neural network through backpropagation. The second model in the proposed system is also a fine-tuning model (different from fine-tuning model of the autoencoders), which is another neural network. A synonym WordNet dataset was used to train the fine-tuning model, where words with the same synset ID are synonyms and represent the same concept. The dataset was scanned to select synonyms and at the same moment five random negative words (not synonyms) were selected per word. The neural network was composed of 150 units in the first hidden layer followed by 100 units in the second hidden layer. The output layer computes cosine similarity between word pairs, where the loss and posterior probability are computed and used to increase similarities between synonyms and decrease similarities with the negative examples. The experiments of the proposed system proved that higher dimensionality improves the similarities between synonyms, and in general the two proposed models conducted relatively better results than word2Vec on synonyms similarities measurement.

---

[2] https://ai.googleblog.com/2018/11/open-sourcing-bert-state-of-art-pre.html

A novel procedure for word vectors fine-tuning with language specific rules was presented by Vulić et al. [11], where two main problems of word embeddings were addressed. The two main problems are learning proper vectors for words with low frequencies and ability to recognize and encode semantically related words e.g. play and plays. This work concentrated on four languages with different inflection levels. The used languages were English, Italian, German, and Russian, where all words with a frequency of more than ten from each language's Wikipedia were extracted. Then this huge set of words was used to generate language specific linguistic constraints through some basic if-else statements, and these generated linguistic constraints were used by an ATTRACT-REPEL model [12]. ATTRACT-REPEL is a framework used for encoding antonymy and similarity information into pretrained word embeddings, so the previously generated linguistic constraints are provided to the framework as collections of ATTRACT and REPEL in addition to the pretrained word embeddings and the framework starts to modify the embeddings to make similar words closer to each other and to push antonyms further apart from each other. The ATTRACT-REPEL algorithm was proposed by Mrkšić et al. [12], so the main contribution of this work was the method of linguistic constraints extraction, which differs from other similar works in avoiding the use of available semantic databases e.g. WordNet [13], or Paraphrase Database [14]. For instance, the authors used two simple rules for English to build ATTRACT constraints from the huge amount of data provided through English Wikipedia, and the rules were:

1. If $w_1$ & $w_2 \in$ English, and $w_1 = w_2 +$ [ing / ed / s] (e.g. w1 = playing and w2 = play), then ($w_1$, $w_2$) & ($w_2$, $w_1$) will be added to ATTRACT constraints collection.

2. If $w_1$ & $w_2 \in$ English, and $w_1$ ends with e & $w_2 = w_1[: -1] + $ (ing / ed), then $(w_1, w_2)$ & $(w_2, w_1)$ will be added to ATTRACT constraints collection.

For antonyms, if $w_1$ & $w_2 \in$ English and w2 can be produced by adding to w1 any member of the following list as prefix: [dis, un, ir, il, in, im, mis, non, anti] (e.g. w1 = relevant and w2 = irrelevant), then $(w_1, w_2)$ & $(w_2, w_1)$ will be added to REPEL constraints collection. The results presented by the authors showed that morph-fitting procedure succeeded in improving pretrained word embeddings, and this fine-tuned word embeddings were able to improve a language understanding models' performance.

**The Proposed Approach**

The main goal of the proposed approach was to improve text classification results by allowing 'class' or 'label' data, to act as contextual factors that contribute to existing pre-trained word embeddings. In other words, the main idea behind the proposed approach was to fine tune a set of word embeddings from an input training dataset such that the values of individual word vectors are conditioned by classes in which they frequently appear. In most existing pre-trained embedding models, the concept of a class does not exist, so taking the terms 'good' and 'bad' for example, it can be observed that the word vectors for those will usually appear close to each other in the vector space, i.e. will be considered semantically similar as both words tend to appear in similar textual contexts. This might not be desirable for certain tasks such as sentiment analysis. By infusing the 'class' to which they belong as an additional contextual factor, the distance between them will increase. So, basically the goal was to take a large set of pre-trained word embeddings in a domain as close as possible to the domain being targeted by a classification task, and use available training data to fine tune those to a very specific task as part of the training process.

To achieve this goal, the work published by Le and Mikolov [5] about representing whole sentences or documents as vectors (Doc2Vec), was used. As stated in the background, Doc2Vec uses a neural net similar to that used for training conventional Word2vec vectors [3], except that it utilizes an additional vector to represent sentences or documents in the training phase and to learn embeddings for these sentences or documents. Basically, the process of learning a document representation is not only conditioned by words appearing within a document, but since a document vector is used to serve as an additional context parameter during training, the process also has an effect on the values of resulting word vectors as the document parameter contributes to the prediction process of each word in the learning iterations. In other words, a document vector act as a memory of the topic of the current document, and this information about the topic has an effect on the resulting word embeddings. The result is that vectors of words that are present more frequently in a document are highly affected by that document.

In the context of this work, classes or categories replace documents as input to the fine-tuning model; in a sense, this could be thought of as 'class2Vec'. This process is illustrated in Figure 1. By training an embeddings model using Doc2Vec, and using the label/class of the document as an additional input in place of a document or paragraph id, words that are indicative of a given class, will have that information encoded in their embeddings.

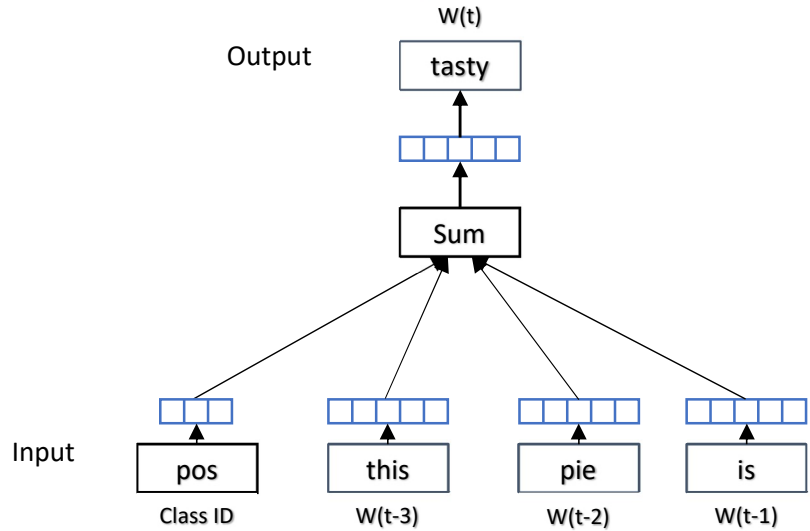

Figure 1: Illustration of how the CBOW like Doc2Vec is adapted in the proposed approach

Consequently, we have the following observations about learning word embeddings with distributed representation of documents (Doc2Vec):

1. If a certain word is present in a class exclusively, its embedding will be highly affected by its class in the training phase.
2. Similarly, if a certain word is present among several classes, the effect of individual classes on the embedding of that word will be diminished.
3. If a word is present in two or more classes, its embedding will be more influenced by the class in which it appeared with a higher frequency.
4. Finally, the learned word embeddings are domain specific, so the word vectors that were trained for a particular domain (e.g. sentiment analysis or emotion detection) may not be suitable for other domains (e.g. question classification).

According to the previous observations, the distributed representation of documents (Doc2Vec) method can be employed to learn word embedding features that are more relevant to the task at hand.

So, in the proposed method, all text fragments (e.g. tweets) related to a specific class (e.g. a given topic, and emotion or sentiment class (love, sadness, fear, etc.)) are labeled with the name of the class to which they belong. Each text fragment is then passed to the Doc2Vec model with its class identifier instead of a document identifier in order to learn the vectors of both words and classes. In this case, only word vectors are used after the training process, while class vectors are discarded. The exact steps for carrying out this process, are formally described as follows:

> Given some input training corpus **T** containing input instances $t_i$ to $t_n$ where **n** is the number of all training instances in **T** and where each instance $t_i$ has a label **c** where **c** ∈ **C** and given a set of pre-trained word embeddings **W** where each word $w_i$ is represented with a vector $wv_i$ of length **m**, the proposed approach can be summarized in the following steps:
>
> 1. Load a pre-trained word embeddings model **W** which has a vocabulary set **V**
> 2. Load training data **T**
> 3. Extract vocabulary set $V_T$ from **T** (this will contain all unique words that appeared in all documents in **T**)
> 4. Identify words that appear in $V_T$ but not **V** (previously unseen words in the pre-trained model) $V_{unseen} = V_T - V$
> 5. For each word $w_i$ in $V_{unseen}$, initialize a random word vector $wv_{unseen\_i}$ of length **m**, and append it to **W**
> 6. Initialize a doc2vec model of vocabulary size $|V \cup V_T|$ with weights from **W** (as updated in the previous step)
> 7. Set the number_of_epochs parameter for training the doc2Vec model to a number that is large enough to fine tune pre-trained embeddings, but which is not too large as to significantly change the pre-trained word vectors. In the experiments presented later in this paper, this parameter was empirically set to 10.
> 8. Use each training instance $t_i$ with class $c_i$, as a training instance for doc2Vec
> 9. Save the resulting embeddings as **W'**. **W'** now contains the fine tuned set of embeddings.

An overview of the proposed approached is illustrated in Figure 2. [3]

---

[3] The code is available at: https://github.com/AmrMehasseb/Embeddings-finetuning

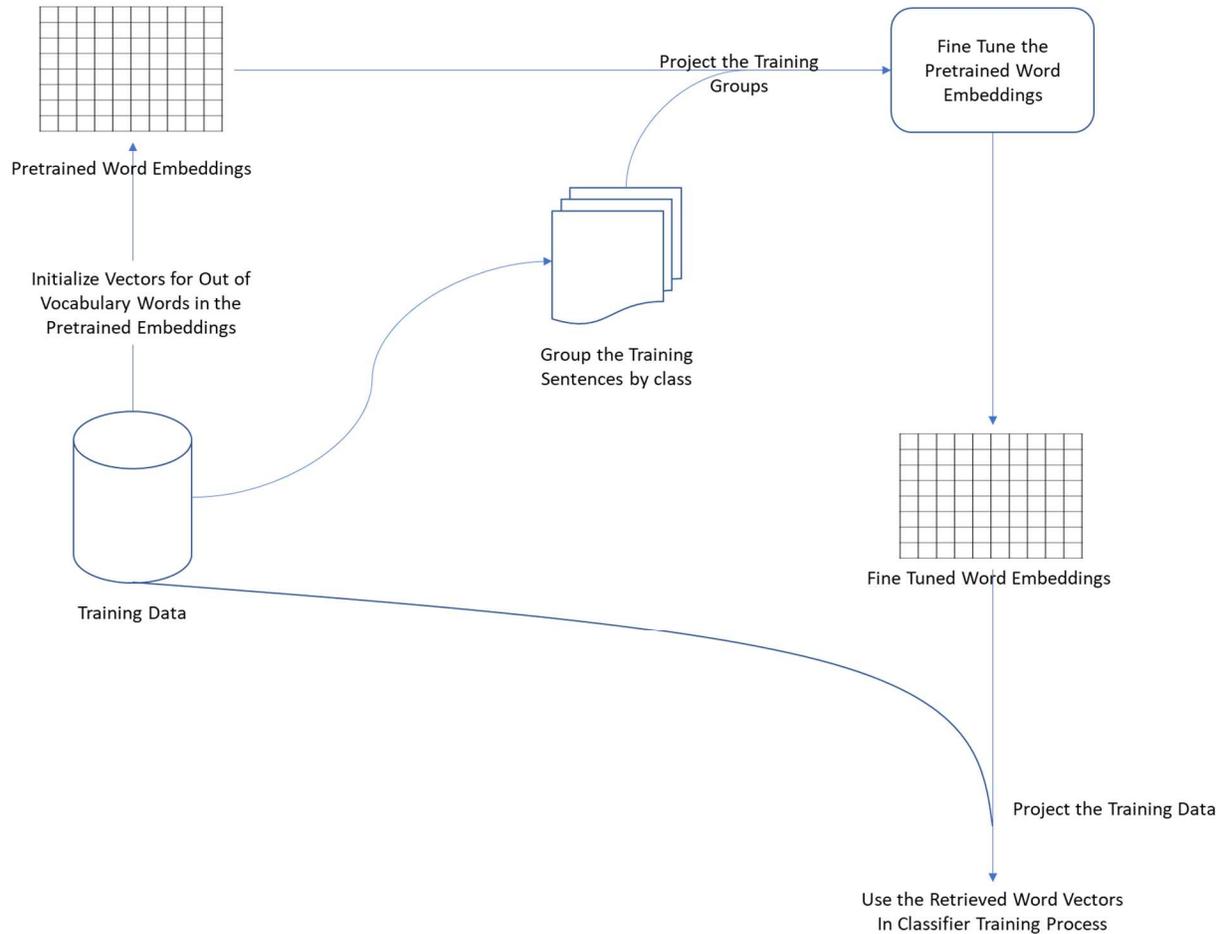

Figure 2: Overview of the proposed approach for fine tuning embeddings

The use of pretrained word vectors for initializing the Doc2Vec model before training it, has the advantage of capturing previously learnt features and only modifying these features so as to reflect their association with classes in the training data. This ensures that words that have not appeared in the training data, due to its limited size, are still likely to have vectors in the final embeddings model. So, during the testing phase (or prediction) words that have not been seen before in the training phase, will still be able to contribute to the overall performance of the system.

**Experiments and Results**

To verify that the previously proposed model has a positive impact on the task of text classification, it was tested it on a variety of benchmarked Arabic and English classification datasets. In these experiments a simple recurrent neural network (RNN) with 64 LSTM units was used [15]. The output layer was a fully connected one with a soft-max function, which produces the probability distribution over labels using equation 1.

$$\text{Soft-Max}_j (x^T w + b) = \frac{e^{x^T w_j + b_j}}{\sum_{k=1}^{K} e^{x^T w_k + b_k}}$$

**Equation 1**

Here **w** and **b** represent the **w**eight and **b**ias vectors respectively.

**Experiment 1: Detecting Emotions in Arabic Text**

In the first set of experiments, an Arabic Twitter Emotions dataset [16] which was annotated for eight emotion classes (sadness, anger, joy, surprise, love, sympathy, fear, and none), was used. This dataset is composed of 10,065 tweets broken down among classes as shown in table 1. In these experiments the simple neural architecture described in the previous section, was employed. Three different experiments were carried out on this dataset: in the first experiment, a baseline was established with random embeddings which were learned from the training data. In the second, AraVec [6], which is set pre-trained set of Arabic word embeddings model, was used. In this specific experiment, the model trained on twitter using skip-gram and having a dimensionality of 300, was used. In the third and last experiment, the AraVec model was fine-tuned using the proposed methodology. The same set of hyperparameters was used across all experiments.

In the original work presenting this dataset, a benchmark of 68.12% in terms of accuracy was reported [16]. This result was obtained after experimenting with a number of traditional machine learning classifiers where a complement Naïve Bayes algorithm (CNB) [17] yielded the best results. This figure is included as a baseline for the reported results. The results of these experiment are shown in table 2. The results show that fine-tuned AraVec significantly improves the results.

**Table 1.** Breakdown of emotions in the final dataset.

| Emotion | Count |
|---|---|
| Sadness | 1256 |
| Anger | 1444 |
| Joy | 1281 |
| Surprise | 1045 |
| Love | 1220 |
| Sympathy | 1062 |
| Fear | 1207 |
| None | 1550 |
| **Total** | **10,065** |

**Table 2.** The baseline results and the results with AraVec embeddings before and after fine-tuning.

|  | Precision weighted average | F-measure weighted average | Overall Accuracy |
|---|---|---|---|
| **Baseline – CNB [16]** | 0.688 | 0.658 | 0.681 |
| **RNN with random embeddings** | 0.641 | 0.635 | 0.643 |
| **RNN with AraVec** | 0.721 | 0.716 | 0.717 |
| **RNN with fine-tuned AraVec** | **0.756** | **0.737** | **0.741** |

**Experiments 2 and 3: Detecting Sentiment in the Stanford sentiment tree dataset**

The second dataset that was used for testing the proposed model, is the Stanford sentiment tree (SST-1) dataset [18]. The dataset is composed of 11,855 sentences divided into 5 classes (very positive, positive, neutral, negative, and very negative). The dataset is provided as 3 data splits (train, test, and dev). Training data was composed of 8,544 snippets while dev data was composed of 1,101 snippets and test data was composed of 2,210 snippets, in addition to the phrases set, which contains the original snippets from SST dataset with their subparts. In this experiment, the previously described RNN model was used and the model was trained using the train split augmented with the subparts from the phrases set, fine-tuned using the dev split, and tested using the test split. The word embeddings model was initialized with pretrained Google news vectors[4] which have a dimensionality of 300. These embedding were fine-tuned using the phrases set provided with the SST (after removing all the testing sentences with their subparts from the set) that contains more than 184,000 labeled phrases. In the carried-out experiments, Google news embeddings were initially used without any fine-tuning. This baseline experiment resulted in an accuracy score of 46.6% total accuracy. The same experiment was repeated using the fine-tuned embeddings. This experiment resulted in accuracy of 49.3%. The overall results (precision, f-score, and accuracy) and a comparison to other published methods are shown in table 3. These results show, that in terms of f-score and accuracy, the proposed method outperforms all the systems cited, despite the simplicity of the architecture employed.

---

[4] https://code.google.com/archive/p/word2vec/

**Table 3.** The results of using Google news embeddings for classification of SST-1 testing set before and after fine-tuning and against other methods.

|  | Precision Weighted Average | F-measure Weighted Average | Overall Accuracy |
|---|---|---|---|
| **RNN with Google news embeddings** | **0.5** | 0.469 | 0.466 |
| **RNN with fine-tuned Google news embeddings** | 0.495 | **0.488** | **0.493** |
| Kim's CNN-non-static [19] | - | - | 0.48 |
| Paragraph-Vec (Le and Micolov) [5] | - | - | 0.487 |
| DCNN (Kalchbrenner) [20] | - | - | 0.485 |
| Dos Santos, Gatti [21] | - | - | 0.483 |

The third dataset on which we experimented on is the also the SST but with the neutral snippets removed and the classes are reduced to positive and negative only (SST-2). A comparison between the results obtained using our proposed method and other published works is presented in table 4. While the presented approach, does result in an improvement compared to the baseline, its overall accuracy is less than that reported in **[19]** and almost equivalent to that of **[5]**. This can be attributed to the simplicity of the architecture employed to test this approach.

**Table 4.** The results of using Google news embeddings for the classification of the SST-2 test set before and after fine-tuning and against other methods.

|  | Precision Weighted Average | F-measure Weighted Average | Overall Accuracy |
|---|---|---|---|
| **RNN with Google news embeddings** | 0.863 | 0.863 | 0.863 |
| **RNN with fine-tuned Google news embeddings** | **0.877** | **0.877** | 0.877 |
| Kim's CNN -multichannel [19] | - | - | **0.881** |
| Paragraph-Vec (Le and Micolov) [5] | - | - | 0.878 |
| DCNN (Kalchbrenner) [20] | - | - | 0.868 |
| Dos Santos, Gatti [21] | - | - | 0.857 |

**Experiment 4: Sentiment Analysis in Arabic Text**

In the fourth experiment, the SemEval 2017 task 4 – subtask A – Arabic dataset [22], was used to evaluate the proposed approach. The dataset is composed of 3355 labeled tweets for training, 671 labeled tweets for validation, and 2499 labeled tweets for testing. The AraVec model was fine-tuned using the training split, and the classification model (which is composed of a simple RNN similar to the previous experiments) was trained on the training split, tuned with the validation split, and tested on the testing set. The baseline experiment results with AraVec were not as good as the top performer model in task 4 – subtask A (NileTMRG [22]), but the proposed methodology achieved an improvement with a 0.621 $F_1$ average score (in SemEval. The results obtained using our method, baseline, and other published works is presented in table 5.

**Table 5.** The results of our method, baseline, and the SemEval 2017 task 4 (subtask-A Arabic) top 3 participants [22].

| Rank | System | R (average recall) | $F_1$ (average f score) | Accuracy |
|---|---|---|---|---|
| - | RNN with fine-tuned AraVec | **0.598** | **0.621** | **0.601** |
| - | RNN with AraVec | 0.547 | 0.535 | 0.576 |
| 1 | NileTMRG | 0.583 | 0.610 | 0.581 |
| 2 | SiTAKA | 0.550 | 0.571 | 0.563 |
| 3 | ELiRF-UPV | 0.478 | 0.467 | 0.508 |

**Experiment 5: SemEval 2018 multilabel classification**

Finally, the proposed method was used for addressing SemEval 2018 task 1 (Affect in Tweets) – subtask E-c (Emotion Classification) in Arabic. AraVec embeddings were used again in this experiment. The provided Arabic dataset was composed of 2,863 training tweets, 1,518 testing tweets, and 585 development tweets. This experiment is different from the previous tasks as targets multilabel classification (the classes are not exclusive, and each tweet can belong to more than one class). The classes used in labeling tweets in this dataset were anger, anticipation, disgust, fear, joy, love, optimism, pessimism, sadness, surprise, and trust. As in the previous experiments, a simple recurrent neural network (RNN) with 64 LSTM units was used in addition to an output layer with 11 units and Sigmoid activation function. In this experiment, using AraVec embeddings before fine-tuning resulted in the highest Jaccard similarity score, which was the score used by the task organizers to rank task participants. The score obtained in this experiment was 50.1 % compared to the best reported score for this task which was 48.9% [23]. Using the fine-tuned embeddings resulting in a score of 49.8 %, which is a better score than the highest reported score, but not quite as high as the result obtained using pre-tuned AraVec. This decrease in performance might be an indication that the proposed method is not suited for multilabel prediction tasks, and that it is more suited for exclusive multiclass text classification tasks. The results of this experiment as well as the reported results of Task 5 emotion classification (E-c) (Arabic sub-task) are presented in table 6.

**Table 6.** The results of our method, baseline, and the SemEval 2018 task 5 E-c Arabic top 3 participants.

|  | Jaccard similarity score | Micro F1 | Macro F1 |
|---|---|---|---|
| **RNN with AraVec** | **0.501** | **0.623** | 0.457 |
| **RNN with fine-tuned AraVec** | 0.498 | 0.617 | 0.451 |
| **SemEval 2018 Task 5 emotion classification Arabic Results [23]** | | | |
| 1-EMA | 0.489 | 0.618 | 0.461 |
| 2-PARTNA | 0.484 | 0.608 | **0.475** |
| 3-Tw-StAR | 0.465 | 0.597 | 0.446 |

**Conclusion**

In this work, a simple, but effective approach for fine-tuning word embeddings using Doc2vec was presented. In this approach, a class identifier is passed to Doc2vec instead of a document identifier. This means that the Doc2vec model treats all text instances belonging to some given class as a single document. When building any word embeddings model, the value of produced word vectors is affected by the context in which individual words appear, but here, for a single word, context is not only defined in terms of the frequency of its appearance in proximity to other words, but also by the frequency of its occurrence with respect to certain classes.

To ensure that the proposed approach is effective across multiple languages, we have tested it on different English and Arabic datasets, using a simple neural classification architecture. When applied on the English Stanford sentiment tree binary and fine-grained datasets, notable improvements over the baselines were observed. Similar results were obtained when using the fine tuned embeddings for classifying an Arabic, single class, emotions dataset. The consistency of improved results over these datasets, suggests that fine tuning pre-trained word embeddings using the proposed approach, is very likely to lead to improved results with respect to single label classification tasks. Applying the proposed approach on an Arabic, multi-label classification task, did not result in similar improvements in performance which suggests that this approach might not be well suited for this task.